\documentclass[10pt, a4paper]{article}
\usepackage{lrec-coling2024} % this is the new style

\usepackage{graphicx}
\usepackage{latexsym}
\usepackage{graphicx}
\usepackage{float}
\usepackage{todonotes}
\usepackage{arydshln}
\usepackage[utf8]{inputenc}
\DeclareUnicodeCharacter{FE0F}{\relax}

\usepackage[T1]{fontenc}
\usepackage{microtype}

\usepackage{inconsolata}
\usepackage{adjustbox}
\usepackage{multirow}
\usepackage{booktabs}
\usepackage{tabularx}
\usepackage{soul}
\usepackage{hyperref}

\title{IT5: Text-to-text Pretraining for \\ Italian Language Understanding and Generation}

\name{Gabriele Sarti, Malvina Nissim}
\address{Center for Language and Cognition (CLCG), University of Groningen \\ \{g.sarti, m.nissim\}@rug.nl}

\abstract{
We introduce IT5, the first family of encoder-decoder transformer models pretrained specifically on Italian. We document and perform a thorough cleaning procedure for a large Italian corpus and use it to pretrain four IT5 model sizes. We then introduce the ItaGen benchmark, which includes a broad range of natural language understanding and generation tasks for Italian, and use it to evaluate the performance of IT5 models and multilingual baselines. We find monolingual IT5 models to provide the best scale-to-performance ratio across tested models, consistently outperforming their multilingual counterparts and setting a new state-of-the-art for Italian language generation.
 \\ \newline  }

\begin{document}

\maketitleabstract

\setlength{\parskip}{0pt}

\section{Introduction}

 The text-to-text paradigm introduced by T5~\citep{t5} has been widely adopted as a simple yet powerful generic transfer learning approach for most language processing tasks~\citep{sanh-etal-2022-multitask,aribandi-etal-2022-ext}. Although the original T5 model was trained exclusively on English data, the same architecture has been extended to a massively multilingual setting covering more than 100 languages by mT5 and ByT5~\citep{byt5, xue-etal-2021-mt5}, following recent advances in the multilingual pre-training of large language models such as mBERT, XLM, XLM-R and mDeBERTa~\citep{devlin-etal-2019-bert,conneau-lample-2019-xlm, conneau-etal-2020-unsupervised,he-etal-2023-debertav3}. Multilingual language models were shown to excel in cross-lingual and low-resource scenarios. Still, multiple studies have highlighted their suboptimal scale-to-performance ratio when compared to monolingual counterparts for language-specific applications in which data are abundant~\citep{nozza-etal-2020-what,rust-etal-2021-good}.

For this reason, monolingual T5 models have recently been pretrained to serve specific language communities, covering languages such as Arabic, Portuguese, Vietnamese and Slovenian~\citep{nagoudi-etal-2021-arat5, carmo-etal-2020-ptt5, phan-etal-2022-vit5, slot5}. These models improve over multilingual baselines on language understanding and generation tasks such as news summarization, headline generation and natural language inference in their respective languages.

In this work, we follow a similar approach to pre-train and evaluate four Italian T5 models of different sizes, which we identify as \textbf{IT5}. In Section~\ref{sec:data}, we present the cleaning procedure for the Italian portion of the mC4 corpus~\citep{xue-etal-2021-mt5} used in pre-training the IT5 models. Section~\ref{sec:experiments} describes multilingual baselines and downstream tasks used to evaluate fine-tuned IT5 models and presents the obtained results. Finally, our findings and future directions are summarized in Section~\ref{sec:conclusion}. We make the following contributions:

\begin{itemize}
    \itemsep=0ex \itemindent=0cm 
    \item We introduce a large-scale cleaned version of the Italian mC4 corpus and use it to pre-train four IT5 models of various dimensions.
    \item We introduce \textsc{ItaGen}, a benchmark for Italian language understanding and generation tasks.
    \item We evaluate IT5 on \textsc{ItaGen}, showing improvements over multilingual baselines and previous state-of-the-art work.
    \item We publicly release all the code, data, and pre-trained/fine-tuned checkpoints for further experimentation by the research community.
\end{itemize}

\noindent To the best of our knowledge, our IT5 models are the first publicly available encoder-decoder models pre-trained exclusively on the Italian language. IT5 constitutes a significant contribution to Italian NLP, as evidenced by its prompt adoption by the research community upon its release~(\citealp{laquatra-cagliero-2023-bartit,papucci-etal-2022-evaluating,mousavi-etal-2023-response} \textit{inter alia}), especially in the context of the latest evaluation campaign of Italian NLP tools~(\citealp{leonardelli-casula-2023-dhfbk, hromei-etal-2023-extremita}). This paper serves as the prime reference for IT5, providing all relevant details regarding training data and parameters, a battery of experiments on a collection of tasks, which can be further used as a reference benchmark, especially for Italian generation, and a discussion of its limitations.\footnote{Resources: \url{https://github.com/gsarti/it5}.}

\section{Related Work}

\subsection{Text-to-text Transfer Transformers}

The Text-to-text Transfer Transformer (T5) model~\citep{t5} adapts the original Transformer architecture proposed by~\citet{vaswani-etal-2017-attention} by reformulating multiple natural language processing tasks into a unified text-to-text format and using them alongside masked span prediction for semi-supervised pre-training. The encoder-decoder architecture of T5 is especially suited for sequence-to-sequence tasks~\citep{sutskever-etal-2014-seq2seq}, which cannot be performed by encoder-only models like BERT~\citep{devlin-etal-2019-bert} and can prove to be challenging for decoder-only models like GPTs~\citep{radford-etal-2019-language,brown-etal-2020-gpt3} due to the lack of explicit conditioning on source context. The same architecture can be easily applied to natural language understanding tasks by using a text-to-text format, making the T5 model highly versatile in most NLP settings.

\subsection{Pre-trained Language Models for Italian}

The high technical expertise and heavy computational resources required for developing state-of-the-art models recently exacerbated inequalities in access to state-of-the-art systems for non-English languages. Despite the good amount of linguistic resources currently available, the Italian NLP community can currently count on a small set of publicly available pre-trained language models based mostly on the BERT architecture -- AlBERTo~\citep{polignano-etal-2019-alberto}, UmBERTo\footnote{\url{https://github.com/musixmatchresearch/umberto}} and GilBERTo\footnote{\url{https://github.com/idb-ita/GilBERTo}} \textit{inter alia}, see \citet{miaschi-etal-2022-probing} for a survey -- and most notably on a single decoder-only model for text generation, GePpeTto~\citep{demattei-etal-2020-geppetto}. Our IT5 models fill the current gap in the availability of sequence-to-sequence models, providing natural choices for monolingual tasks such as summarization, question answering and reformulation.

\section{Data and Model Pretraining}
\label{sec:data}

The original T5 models were pre-trained on the 750GB web-scraped English C4 corpus~\citep{t5}. C4 authors cleaned the corpus with heuristics to remove templated fillers, text deduplication, Javascript code, slurs and non-English texts. The multilingual counterpart of T5 adopts a similar procedure to create mC4~\citep{xue-etal-2021-mt5}, a multilingual version of C4 including 107 languages. While mC4 authors adopted a similar procedure, the language detection threshold is lowered to 70\% and other useful heuristics are omitted due to their brittleness across various character systems. As a result, the resulting corpus has an overall lower quality, with recent work finding 16\% of sampled mC4 examples having the wrong language tag, and 11\% not containing any linguistic information~\citep{caswell-etal-2022-quality}. For this reason, we perform a thorough cleaning of the Italian portion of mC4 before pre-training IT5.

\subsection{Cleaning the Italian mC4 Corpus}

The original Italian mC4 Corpus includes approximately 359GB of raw text data and is one of the largest public Italian corpora. To perform a more thorough cleaning of this data, we use a public implementation\footnote{\url{https://gitlab.com/yhavinga/c4nlpreproc}} reproducing and improving the original C4 data cleaning pipeline. Specifically, we sentence-tokenize documents and remove sentences containing either (i) words from a manually selected subset of the Italian and English List of Dirty Naughty Obscene and Otherwise Bad Words;\footnote{\url{https://github.com/LDNOOBW}} (ii) less than three words, or a word longer than 1000 characters; (iii) an end symbol not matching standard end-of-sentence punctuation for Italian; or (iv) strings associated to Javascript code, lorem ipsum, English and Italian privacy policy/cookie disclaimers. We finally keep only documents containing more than five sentences, having between 500 and 50k characters, and having Italian as the main language.\footnote{We use the \href{https://github.com/Mimino666/langdetect}{langdetect} toolkit.} The resulting Clean Italian mC4 Corpus\footnote{\url{https://hf.co/datasets/gsarti/clean_mc4_it}}, contains roughly 215GB of raw Italian text, corresponding roughly to 103M documents and 41B words.

\subsection{Model and Training Parameters}

The first 10M documents sampled from the cleaned corpus are used to train a SentencePiece unigram subword tokenizer~\cite{kudo-2018-subword} with a vocabulary size of 32k words. The full cleaned corpus is then used to pre-train three models following the canonical small, base and large sizes~\citep{t5}. Moreover, a fourth model adopting the efficient small EL32 architecture by~\citet{tay-etal-2022-scale} is also pre-trained and evaluated.%CAMREADY The overall parametrization and additional training details are presented in Appendix~\ref{app:full-params}.

\begin{table}
\adjustbox{width=\linewidth}{
\begin{tabular}{ll}
\toprule
\textbf{Task} & \textbf{Dataset} \\
\midrule
Wiki Summarization & WITS~\citep{casola-lavelli-2021-wits} \\
News Summarization & NewsSum-IT~\citep{landro-etal-2022-two} \\
\midrule
Question Answering & \multirow{2}{5em}{SQuAD-IT}~\citep{croce-etal-2018-neural} \\
Question Generation & \\
\midrule
Headline Style Transfer & \multirow{2}{6em}{CHANGE-IT}~\citep{demattei-etal-2020-changeit} \\
Headline Generation & \\
\midrule
Formality Style Transfer & XFORMAL-IT~\citep{briakou-etal-2021-ola} \\
\bottomrule
\end{tabular}
}
\caption{Summary of datasets composing \textsc{ItaGen}.}
\label{tab:tasks}
\end{table} % Params

\section{Evaluation}
\label{sec:experiments}

\subsection{The ItaGen Benchmark}

We propose a selection of seven representative tasks, collectively referred to as \textsc{ItaGen}, to evaluate the downstream performances of fine-tuned IT5 and mT5 models. \textsc{ItaGen} aims to provide a comprehensive overview of canonical conditional text generation applications such as summarization, style transfer and question generation, and is constrained by the limited availability of Italian corpora for such tasks. Moreover, we also include a direct comparison of IT5 performances against encoder-based extractive systems for extractive question answering. Table~\ref{tab:tasks} provides an overview of \textsc{ItaGen} tasks and datasets.
\vspace{-7pt}
\paragraph{Wikipedia Summarization} We evaluate encyclopedic summarization on the Wikipedia for Italian Text Summarization (WITS) corpus~\citep{casola-lavelli-2021-wits}, containing 700k articles extracted from a cleaned dump of the Italian Wikipedia alongside their leading sections used as approximated summaries. We adopt the original evaluation setup using a 10k examples test set.
\vspace{-7pt}
\paragraph{News Summarization} We evaluate news article summarization by concatenating Fanpage.it and IlPost newspapers articles collected by~\citet{landro-etal-2022-two}. We refer to this concatenated corpus as NewsSum-IT. We fine-tune our systems on the training set, including roughly 100k articles and respective short summaries and evaluate them separately on the two test sets defined by the dataset creators. We report the averaged metrics across the two newspapers in the results section.
\vspace{-7pt}
\paragraph{Question Answering} We evaluate extractive question answering using the SQuAD-IT dataset~\citep{croce-etal-2018-neural}, containing 50k paragraph-question-answers triplets automatically translated from the English SQuAD dataset~\citep{rajpurkar-etal-2016-squad}. We frame the QA task as a text-to-text problem aimed at generating responses given a source text using the format \texttt{<CONTEXT> Domanda: <QUESTION>}. We use the original evaluation script and splits.
\vspace{-7pt}
\paragraph{Question Generation} We use SQuAD-IT to evaluate question generation capabilities by reordering the text triplets, making the model predict a plausible question given a source text in the format \texttt{<CONTEXT> Risposta: <ANSWER>}, where the answer is the first among the available per-example answers, using the same train-test splits of QA.
\vspace{-7pt}
\paragraph{Headline Style Transfer} We evaluate style transfer abilities in the news domain on the CHANGE-IT shared task~\citep{demattei-etal-2020-changeit}, containing 60k newspaper articles and headlines from the left-leaning Italian newspaper la Repubblica and the right-leaning Il Giornale, respectively. We train and validate our models on author-defined splits using the original cross-source article-to-headline generation. %since we find it to work more effectively in absence of high-quality topical alignment of articles from the two newspapers.
We report average scores for the two style transfer directions (Il Giornale $\leftrightarrow$ la Repubblica).
\vspace{-7pt}
\paragraph{Headline Generation} To evaluate news headline generation, we combine the two CHANGE-IT subsets to create a corpus of roughly 120k news articles and headlines pairs, which we refer to with the name HeadGen-IT. Original CHANGE-IT test sets are preserved.
\vspace{-7pt}
\paragraph{Formality Style Transfer} We evaluate the formality style transfer capabilities of our models on the Italian subset of the XFORMAL dataset~\citep{briakou-etal-2021-ola}, containing a training set of ~115k forum messages automatically translated from the GYAFC corpus~\citep{rao-tetreault-2018-dear} and covering the topics of entertainment, music, family and relationships, and a small test set of 1000 formal-informal pairs obtained directly in Italian from four crowd workers via Amazon Mechanical Turk. We evaluate our models in both style transfer directions (Formal $\leftrightarrow$ Informal).

\subsection{Evaluation Metrics}

We use a combination of common lexical and trainable metrics across all available tasks. We use the language-independent ROUGE metric~\citep{lin-2004-rouge} in its R1, R2 and RL variants to evaluate lexical matches, and BERTScore~(\citeauthor{zhang-etal-2020-bertscore} \citeyear{zhang-etal-2020-bertscore}; BS) to evaluate correspondence at the semantic level.\footnote{We use an Italian BERT model to obtain baseline scores to broaden the metric range and remove noise, following authors' recommendations.%CAMREADY Details in Appendix~\ref{app:bertscore-baseline}.
} For QA, the canonical exact-match (EM) and F1-score (F1) metrics are used. Finally, for the news headline style transfer task, we use trained classifiers provided by the authors\footnote{\href{https://github.com/michelecafagna26/CHANGE-IT}{\texttt{michelecafagna26/CHANGE-IT}}} to ensure comparable headline-headline (HH) and headline-article (HA) coherence performances.

\subsection{Baselines}

Besides baselines available from previous studies using the selected datasets, we also adopt the same fine-tuning procedure for fine-tuning two sizes (small and base) of the multilingual T5 model (mT5)~\cite{xue-etal-2021-mt5}. These multilingual models are used to assess the validity of our pre-training procedure and to observe whether the monolingual setting improves performance.\footnote{mT5 models are bigger than T5s due to larger vocabularies and embedding matrices, making their usage on consumer accelerators more challenging.} %CAMREADYAdditional details on the parametrization across fine-tuning tasks for mT5 and IT5 models are provided in Appendix~\ref{app:ft-params}.

\begin{table*}
\centering
\adjustbox{width=\textwidth}{
\begin{tabular}{ccc}

\begin{tabular}[t]{l|cccc}
\toprule
& \multicolumn{4}{c}{\textbf{WITS}} \\
\cmidrule(lr){2-5}
 & R1 & R2 & RL & BS \\
\midrule
TextRank~\citeyearpar{casola-lavelli-2021-wits}    & .302 & .076 & .197 & - \\
LexRank~\citeyearpar{casola-lavelli-2021-wits}     & .269 & .059 & .175 & - \\
SumBasic~\citeyearpar{casola-lavelli-2021-wits}    & .206 & .048 & .140 & - \\
%IT5 Small~\citeyearpar{casola-lavelli-2021-wits}  & .216 & .097 & .193 & - \\
\midrule
mT5 Small & .347 & .200 & .316 & .517 \\
mT5 Base & .348 & .200 & .315 & .520 \\
\midrule
IT5 Small (ours) & .337 & .191 & .306 & .504 \\
IT5 EL32 (ours)  & .346 & .196 & .314 & .513 \\
IT5 Base (ours)  & \textbf{.369} & \textbf{.217} & \textbf{.333} & \textbf{.530} \\
IT5 Large (ours) & .335 & .191 & .301 & .508 \\
\bottomrule
\end{tabular}

&

\begin{tabular}[t]{l|cccc}
\toprule
& \multicolumn{4}{c}{\textbf{CHANGE-IT}} \\
\cmidrule(lr){2-5}
 & HH & HA & RL & BS \\
\midrule
PointerNet~\citeyearpar{demattei-etal-2020-changeit} & .644 & .874 & .151 & - \\
BiLSTM$_{\texttt{+Att}}$~\citeyearpar{demattei-etal-2020-interaction} & .744 & .846 & .155 & - \\
\midrule
mT5 Small & .777 & .807 & .211 & .372 \\
mT5 Base &  .795 & .799 & .236 & .398 \\
\midrule
IT5 Small (ours) & .898 & \textbf{.882} & .231 & .392 \\
IT5 EL32 (ours)  & .822 & .786 & .244 & .406 \\
IT5 Base (ours) &  \textbf{.904} & .868 & \textbf{.247} & \textbf{.411} \\
IT5 Large (ours) & .895 & .861 & .237 & .390 \\
\bottomrule
\end{tabular}

&

\begin{tabular}[t]{l|c|cc}
\toprule
& \textbf{Size} & \multicolumn{2}{c}{\textbf{SQuAD-IT QA}} \\
\cmidrule(lr){3-4}
 & \# & F1 & EM \\
 \midrule
DrQA-IT~\citeyearpar{croce-etal-2018-neural}                         &    - & .659 & .561          \\
mBERT~\citeyearpar{croce-etal-2019-deep}                             & 110M & .760 & .650          \\
BERT-IT~\footnotemark~\citeyearpar{devlin-etal-2019-bert}               & 110M & .753 & .638          \\
%MiniLM~\citeyearpar{riabi-etal-2021-synthetic}                       &  66M & .720 & .577          \\
%MiniLM$_{\texttt{+st}}$~\citeyearpar{riabi-etal-2021-synthetic}      &  66M & .745 & .620          \\
XLM-R Large$_{\texttt{+st}}$~\citeyearpar{riabi-etal-2021-synthetic} & 560M & \textbf{.804} & .676 \\
\midrule
mT5 Small & 300M & .660 & .560 \\
mT5 Base & 580M  & .757 & .663 \\
\midrule
IT5 Small (ours) & 60M  & .716 & .619 \\
IT5 EL32 (ours) & 143M & .747 & .645 \\
IT5 Base (ours) & 220M  & .761 & .663 \\
IT5 Large (ours) & 738M & .780 & \textbf{.691} \\
\bottomrule
\end{tabular}
\vspace{5pt}

\\

\multicolumn{3}{c}{
\begin{tabular}[b]{l|cccc|cccc|cccc|cccc|cccc}
\toprule
& \multicolumn{4}{c}{\textbf{NewsSum-IT}} & \multicolumn{4}{c}{\textbf{SQuAD-IT QG}} & \multicolumn{4}{c}{\textbf{HeadGen-IT}} & \multicolumn{4}{c}{\textbf{XFORMAL-IT F$\rightarrow$I}} & \multicolumn{4}{c}{\textbf{XFORMAL-IT I$\rightarrow$F}} \\
\cmidrule(lr){2-5}
\cmidrule(lr){6-9}
\cmidrule(lr){10-13}
\cmidrule(lr){14-17}
\cmidrule(lr){18-21}
 & R1 & R2 & RL & BS & R1 & R2 & RL & BS & R1 & R2 & RL & BS & R1 & R2 & RL & BS & R1 & R2 & RL & BS \\
\midrule
mBART Large~\citeyearpar{landro-etal-2022-two} 
            & .323    & .150    & .248    & -       & -       & -       & -       & -       & -       & -       & -       & -       & -       & -       & -       & -       & -       & -       & -       & -       \\
\midrule
mT5 Small   & .340    & .161    & .262    & .375    & .306    & .143    & .286    & .463    &.277     & .094    & .244    & .408    & .651    & \bf.450 & .631    & .666    & .638    & .446    & .620    & .684    \\     
mT5 Base    & .330    & .155    & .258    & .393    & .346    & .174    & .324    & .495    &.302     & .109    & .265    & .427    & \bf.653 & .449    & \bf.632 & .667    & .661    & .471    & .642    & .712    \\
\midrule
IT5 Small (ours)  & .354    & .172    & .278    & .386    & .367    & .189    & .344    & .505    &.287     & .100    & .253    & .414    & .650    & \bf.450 & .631    & .663    & .646    & .451    & .628    & .702    \\
IT5 EL32 (ours)   & .339    & .160    & .263    & \bf.410 & .382    & .201    & .357    & .517    &.299     & .108    & .264    & .427    & .459    & .244    & .435    & \bf.739 & .430    & .221    & .408    & .630    \\
IT5 Base (ours) & .251    & .101    & .195    & .315    & .382    & .199    & .354    & .516    & \bf.310 & .112    & \bf.270 & \bf.433 & .652    & .446    & .632    & .665    & .583    & .403    & .561    & .641    \\
IT5 Large (ours) & \bf.377 & \bf.194 & \bf.291 & -       & \bf.383 & \bf.204 & \bf.360 & \bf.522 & .308    & \bf.113 & \bf.270 & .430    & .611    & .409    & .586    & .613    & \bf.663 & \bf.477 & \bf.645 & \bf.714 \\
\bottomrule
\end{tabular}                            
}
\end{tabular}
}
\caption{IT5, mT5 and baseline models performances on \textsc{ItaGen} datasets. Best scores are highlighted.}
\label{tab:results}
\end{table*}

\footnotetext[11]{\href{https://huggingface.co/antoniocappiello/bert-base-italian-uncased-squad-it}{\texttt{bert-base-italian-uncased-squad-it}}}

\subsection{Results and Discussion}
\label{sec:results}

Table~\ref{tab:results} present the results of our fine-tuning experiments. Given the broad scope of our analysis, we limit ourselves to comment on salient trends we observe across tasks and model categories.
\vspace{-7pt}
\paragraph{IT5 models provide state-of-the-art performances for language generation and understanding tasks in Italian.} The IT5 models outperform multilingual models and previous systems in 6 out of 8 evaluated tasks, with noticeable improvements over mT5 systems, particularly for question answering and generation and for headline-headline coherence on the news headline style transfer task. For QA, the IT5 Large model outperforms most extractive systems, including the XLM-R Large by~\citet{riabi-etal-2021-synthetic}, despite its ad-hoc synthetic data augmentation procedure.
\vspace{-7pt}
\paragraph{Multilingual models can still be helpful in specific applications and when using translated data.} We observe that multilingual language models perform best in the news summarization and the formal-to-informal style transfer tasks. In the case of news summarization, we attribute the performance gap in large part to the scale of the mBART baseline model. For the formality style transfer task, after a preliminary error analysis, we conjecture that translation errors and English acronyms present in the noisy training split of XFORMAL act as out-of-distribution samples in the monolingual setting, disrupting the performances of IT5 systems but are captured more easily by multilingual systems which were exposed by multiple data distributions by design. This would indicate a better fit of multilingual pre-trained models for such settings if verified. We leave a more thorough analysis of these patterns to future work.
\vspace{-7pt}
\paragraph{Scaling model size does not guarantee an increase in performance if not supported by an increase in computational resources.} Contrary to common scaling trends for Transformers~\cite{brown-etal-2020-gpt3}, we do not observe a systematic increase in downstream performances for IT5 models when increasing their size, despite lower loss scores and higher accuracies achieved by larger models during pre-training. While recent work highlighted how better pre-training performances do not always correspond to better downstream scores for T5 models~\citep{tay-etal-2022-scale}, we hypothesize that our results might be related to a bottleneck in the maximal batch size for large models, set to 128 examples instead of the 2048 reported by~\citet{t5} due to lack of resources. We observe that the EL32 architecture can frequently outperform larger model variants, suggesting efficient model design as a promising direction for monolingual model development.

\section{Conclusion}
\label{sec:conclusion}

This paper introduced IT5, the first family of large-scale encoder-decoder models pre-trained in Italian. We presented a detailed overview of the overall training and evaluation procedure, including comparisons with multilingual counterparts on a broad set of Italian language generation tasks. We obtained new state-of-the-art results across most evaluated tasks and concluded by discussing the shortcomings of large-scale monolingual language modeling when dealing with automatically translated data and limited computational resources.

In light of our results, we deem a further investigation of time and quality trade-offs between pre-trained monolingual models and a language-specific continued pre-training of multilingual models as a future step to further narrow the gap in modeling performances for less-resourced languages.

\section{Acknowledgements}

We thank the Google TPU Research Cloud program for providing us with free access to TPU v3-8 machines used in pre-training the IT5 models and the Center for Information Technology of the University of Groningen for providing access to the Peregrine high-performance computing cluster used in fine-tuning and evaluation experiments. We also thank the Huggingface team for creating the original template for TPU-compatible pre-training and fine-tuning scripts for T5 models and making them available during the Huggingface JAX/Flax Community Week.

\section{Ethics Statement}

Despite our thorough cleaning procedure aimed at removing vulgarity and profanity, it must be acknowledged that models trained on web-scraped contents such as IT5 will inevitably reflect and amplify biases present in Internet blog articles and comments, resulting in potentially harmful content such as racial or gender stereotypes and conspiracist views. In light of this, we encourage further studies to assess the magnitude and prevalence of such biases. Model usage should ideally be restricted to research-oriented and non-user-facing endeavors.

Due to our limited computational resources, we could not conduct an exhaustive hyperparameter search for pre-training and fine-tuning IT5 models. For this reason, reported scores should not be treated as the best achievable results, and further improvements can undoubtedly be achieved with additional benchmarking effort.

Finally, despite using standard metrics capturing lexical and semantic similarity to evaluate our models, we do not explicitly evaluate the factual consistency of generated outputs. For this reason, the real-world effectiveness of our models should be further assessed in future studies, especially for tasks prone to hallucination, such as abstractive summarization.

%\nocite{*}
\section{Bibliographical References}\label{reference}
\label{main:ref}

\bibliographystyle{lrec-coling2024}
\bibliography{lrec-coling2024}

%\section{Language Resource References}
%\label{lr:ref}
%\bibliographystylelanguageresource{lrec-coling2024-natbib}
%\bibliographylanguageresource{languageresource}

\appendix
\newpage

\section{Model Parametrization and Additional Pretraining Details}
\label{app:full-params}

\begin{figure}
    \includegraphics[width=0.5\textwidth, angle=0]{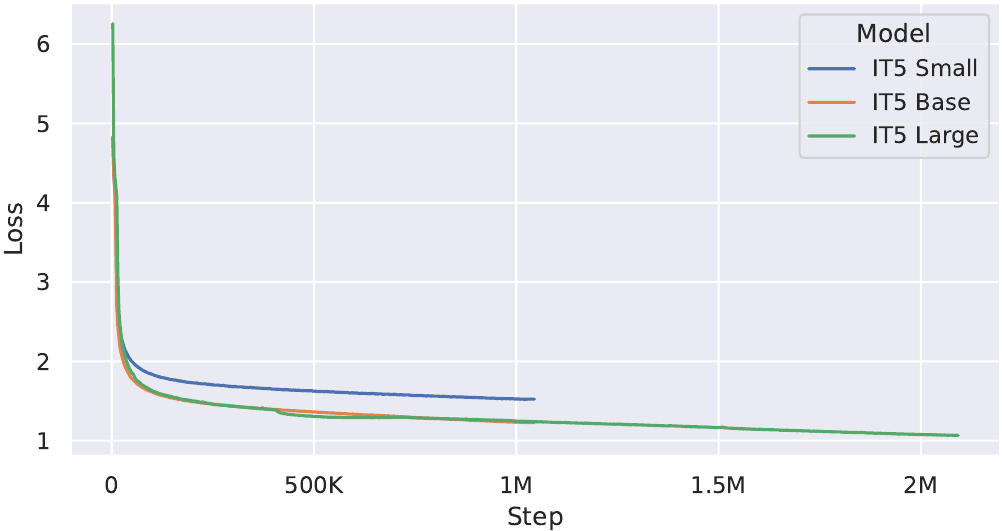}
    \caption{Loss curves for the masked span prediction task used to pre-train the IT5 models.}
    \label{fig:learning-curves}
\end{figure}

Models are trained on a TPU v3-8 accelerator on Google Cloud Platform using the JAX framework~\citep{jax2018github} and Huggingface Transformers~\citep{wolf-etal-2020-transformers}. We adopt the T5 v1.1 architecture\footnote{\href{https://github.com/google-research/text-to-text-transfer-transformer/blob/main/released_checkpoints.md\#t511}{\texttt{text-to-text-transfer-transformer/t511}}} also used by the mT5 model, improving upon the original T5 by using GeGLU nonlinearities~\citep{shazeer-2020-glu}, scaling model hidden size alongside feedforward layers and pre-training only on unlabeled data, without dropout. All models are pre-trained with a learning rate of 5e-3 and a maximum sequence length of 512 tokens using the Adafactor optimizer~\cite{pmlr-v80-shazeer18a} to reduce the memory footprint of training and are validated on a fixed subset of 15'000 examples. Figure~\ref{fig:learning-curves} shows the computed loss during the training process for the three standard models (excluding the efficient one). We used the Google Cloud Carbon Footprint tool to estimate the overall amount of CO2 generated by the pre-training process and found it to be approximately equal to 7kgCO2, corresponding approximately to the emissions of a 60km car ride.\footnote{\url{https://ec.europa.eu/eurostat/cache/metadata/en/sdg\_12\_30\_esmsip2.htm}}

\begin{table}
\begin{small}
\begin{center}
\begin{tabular}{lccc}
\toprule
 & \textbf{Small} & \textbf{Base} & \textbf{Large} \\
\midrule
\# of parameters & 60M & 220M & 738M \\
\# of steps & 1.05M & 1.05M & 2.1M \\
Training time & 36 h & 101 h & 370 h \\
Batch size & 128 & 128 & 64 \\
Weight decay & 1e-3 & 1e-3 & 1e-2 \\
Feedforward size & 1024 & 2048 & 2816 \\
Hidden size & 512 & 768 & 1024 \\
\# encoder layers & 6 & 12 & 24 \\
\# decoder layers & 6 & 12 & 24 \\
\# attention heads & 6 & 12 & 16 \\
\midrule
K-V proj. size & \multicolumn{3}{c}{64} \\
Dropout rate & \multicolumn{3}{c}{0} \\
Non-linearity & \multicolumn{3}{c}{Gated GeLU} \\
LayerNorm $\epsilon$ &  \multicolumn{3}{c}{1e-6} \\
\# rel. att. buckets &  \multicolumn{3}{c}{32} \\
Vocabulary size & \multicolumn{3}{c}{32'000}\\
\bottomrule
\end{tabular}
\end{center}
\end{small}
\caption{Full parametrization for IT5 models. Parameters below the line are shared across all configurations.}
\label{tab:full-params}
\end{table} % Full Params

Table~\ref{tab:full-params} shows the full parameter configuration for all IT5 model sizes. The models correspond to the three canonical sizes for T5 models (Small, Base, Large) with T5 v1.1 improvements, and the efficient small version with 32 encoder layers introduced by~\citet{tay-etal-2022-scale} (EL32).

\section{Italian Baseline Scores for BERTScore Rescaling}
\label{app:bertscore-baseline}

Table ~\ref{tab:bertscore-rescale} contains the baseline scores computed on the first 1M examples of the Cleaned Italian mC4 Corpus using the same model, which we later use for evaluating generation performances. These should be used alongside the same model and the \texttt{--rescale\_with\_baseline} option to obtain BERTScore performances directly comparable to the ones reported in this work. 

The hash code used for reproducibility by the BERTScore library is \texttt{dbmdz/bert-base-italian-xxl-uncase\\d\_L10\_no-idf\_version=0.3.11(hug\_tr\\ans=4.16.0)-rescaled}

\section{Parametrization for Fine-tuning Experiments}
\label{app:ft-params}

Table~\ref{tab:ft-task-params} contains task-specific parameters that were used for the fine-tuning experiments. For mT5 Small, IT5 Small and IT5 Base models we use a learning rate of 5e-4 and a batch size of 64 examples, while larger models (mT5 Base and IT5 Large) were fine-tuned with a leaning rate of 5e-5 and a batch size of 32. All models are fine-tuned with linear schedule with no warmup using the AdamW optimizer~\citep{loshchilov-hutter-2018-decoupled}.

We highlight that the batch sizes used for fine-tuning are significantly smaller from the canonical batch size of 128 adopted by~\citet{t5} due to hardware limitations.

\begin{table}
\begin{center}
\begin{small}
\begin{tabular}{lccc}
\toprule
\textbf{Layer} & \textbf{Precision} & \textbf{Recall} & \textbf{F1} \\
\midrule
0  & 0.3164 & 0.3165 & 0.3100 \\
1  & 0.3869 & 0.3870 & 0.3843 \\
2  & 0.3777 & 0.3778 & 0.3759 \\
3  & 0.4955 & 0.4955 & 0.4945 \\
4  & 0.5646 & 0.5646 & 0.5637 \\
5  & 0.5874 & 0.5874 & 0.5868 \\
6  & 0.5712 & 0.5713 & 0.5706 \\
7  & 0.5483 & 0.5484 & 0.5478 \\
8  & 0.4989 & 0.4989 & 0.4979 \\
9  & 0.4401 & 0.4401 & 0.4382 \\
10 & 0.4082 & 0.4082 & 0.4061 \\
11 & 0.3766 & 0.3766 & 0.3750 \\
12 & 0.3400 & 0.3400 & 0.3381 \\
\bottomrule
\end{tabular}
\end{small}
\end{center}
\caption{Baseline scores for using \texttt{dbmdz/bert-base-italian-xxl-uncased} with the BERTScore evaluation framework.}
\label{tab:bertscore-rescale}
\end{table} % Baseline

\begin{table}
\begin{small}
\begin{center}
\begin{tabular}{lccc}
\toprule
\textbf{Dataset} & \textbf{SL} & \textbf{TL} & \textbf{\# Epochs} \\
\midrule
WITS & 100 & 128 & 3 \\
NewsSum-IT & 512 & 128 & 7\\
SQuAD-IT QA & 512 & 64 & 7\\
SQuAD-IT QG & 512 & 128 & 7\\
XFORMAL F $\leftrightarrow$ I & 128 & 64 & 10\\
CHANGE-IT & 512 & 64 & 10\\
HeadGen-IT & 512 & 64 & 7\\
\bottomrule
\end{tabular}
\end{center}
\end{small}
\caption{Task-specific fine-tuning parameters. SL = Max. source length. TL = Max. target length.}
\label{tab:ft-task-params}
\end{table} % Full Params

\section{Generation Examples using IT5 Base}
\label{sec:examples}

Tables~\ref{tab:examples-1} and~\ref{tab:examples-2} present some generation examples for the IT5 Base model across all the evaluated tasks. We use [...] to omit portions of long sources that we judge to be less salient to improve the readability of the examples. Outputs are lowercase because the IT5 Base tokenizer is uncased, while using the EL32 model would produce results with normal casing. Examples shown were randomly sampled among model generations for the respective test sets.

While the quality is generally high, we observe that summarization results, especially for the WITS dataset, tend to contain hallucinated information obtained by combining unrelated portions of the source. For example, "Libro Entertainment" in the first example appears to be a translated version of the actual name of the publishing house, and Paolo Villaggio published an audiobook with the company rather than owning it, as it is stated in the generated summary. This is a well-known problem of abstractive summarization systems~\citep{maynez-etal-2020-faithfulness,ji-etal-2022-survey}, which hasn't been studied extensively for languages other than English.

\begin{table*}[!t]
\small 
\centering
\vspace{-2mm}
\scalebox{0.80}{
    \def\arraystretch{1.5}
    \begin{tabular}{lp{4em}p{46em}}
    \toprule
    \bf{Task} & \bf{Field} & \bf{Examples} \\
    \midrule
    \multirow{3}{8em}{\bf{Wikipedia \\ Summarization \\ (WITS)}} & ️\multirow{3}{4em}{Wikipedia \\ Article \\ (Source)} & Volume Entertainment ha prodotto e pubblicato l'audiolibro ''Fantozzi'', dal primo libro della saga di Paolo Villaggio del 1971, aggiornato e registrato dallo stesso Villaggio nel novembre 2015. [...] Nel 2017 Volume Entertainment produce il docu-film ''La voce di Fantozzi'', riconosciuta opera di interesse culturale dal MIBACT e presentato in concorso in anteprima alla 74 Mostra internazionale d'arte cinematografica di Venezia nello stesso anno. Volume Audiobooks è la divisione dedicata alla produzione e distribuzione di audiolibri e podcast. [...] Volume Media è specializzata nella produzione di documentari e format TV. [...] \\
    \cdashline{2-3}
    & \multirow{2}{4em}{Summary \\ (Generated)} & libro entertainment è una casa editrice italiana fondata nel 2015 da paolo villaggio e l'editore e distributore di audiolibri e podcast.\\
    \cmidrule(lr){2-3}
    & \multirow{3}{4em}{Wikipedia \\ Article \\ (Source)} & Intorno alle 7:15 del mattino del 10 aprile 2002, un militante palestinese si fece esplodere sull'autobus Egged numero 960 che era in viaggio da Haifa a Gerusalemme. [...]  Otto persone furono uccise nell'attacco e altre 19 rimasero ferite, tre delle quali gravemente. La maggior parte dei feriti erano soldati e poliziotti israeliani. La forza dell'esplosione danneggiò anche molti veicoli nelle vicinanze. [...] \\
    \cdashline{2-3}
    & \multirow{2}{4em}{Summary \\ (Generated)} & \multirow{2}{*}{l'attentato di haifa è stato un attacco terroristico israeliano avvenuto il 10 aprile 2002 ad haifa, in israele.} \\
    \\
    \midrule
    \multirow{3}{8em}{\bf{News \\ Summarization \\ (NewsSum-IT)}} & ️\multirow{3}{4em}{News \\ Article \\ (Source)} & Mercoledì 18 settembre il cofondatore e CEO di Google, Larry Page, ha annunciato la creazione di Calico, una nuova società che si occuperà dello sviluppo di soluzioni per la salute e per il benessere. Calico si occuperà principalmente di affrontare i problemi legati all’invecchiamento e ad alcune malattie, come il cancro. [...] Calico sarà principalmente dedicata alla ricerca di cure e terapie contro l’invecchiamento e le malattie degenerative. I piani, come ammette lo stesso Page, non sono ancora del tutto chiari perché il progetto è appena partito e ha obiettivi i cui risultati saranno visibili solo nel lungo periodo. \\
    \cdashline{2-3}
    & \multirow{2}{4em}{Summary \\ (Generated)} & \multirow{2}{*}{google ha creato calico, una nuova società che si occuperà di curare il cancro.} \\
    \\
    \cmidrule(lr){2-3}
    & \multirow{3}{4em}{News \\ Article \\ (Source)} & Clubhouse, un nuovo social network in cui invece di scrivere brevi messaggi o condividere immagini si parla in diretta, sta attirando velocemente le attenzioni di giornali e appassionati di internet anche in Italia, dopo che nelle scorse settimane lo aveva fatto negli Stati Uniti. [...] Esiste dallo scorso marzo, ed è una specie di forum, ma orale invece che scritto. Potrebbe assomigliare a Telegram, se Telegram fosse fatto di soli messaggi vocali, con la differenza che in Clubhouse non c’è nulla di registrato: si parla live, chiedendo di intervenire con una simbolica alzata di mano. [...] \\
    \cdashline{2-3}
    & \multirow{2}{4em}{Summary \\ (Generated)} & il nuovo social network che parla in diretta. è un forum orale, ma orale invece che scritto, e sta attirando le attenzioni di giornali e appassionati di internet. \\
    \midrule
    \multirow{3}{8em}{\bf{Question \\ Answering \\ (SQuAD-IT QA)}} & ️\multirow{3}{4em}{Context + \\ Question \\ (Source)} & La crisi petrolifera del 1973 iniziò nell' ottobre 1973 quando i membri dell' Organizzazione dei Paesi esportatori di petrolio arabo (OAPEC, composta dai membri arabi dell' OPEC più Egitto e Siria) proclamarono un embargo petrolifero. [...] Più tardi fu chiamato il "primo shock petrolifero", seguito dalla crisi petrolifera del 1979, definita il "secondo shock petrolifero". \textit{Domanda: Chi ha proclamato l' embargo petrolifero?} \\
    \cdashline{2-3}
    & \multirow{2}{4em}{Answer \\ (Generated)} & \multirow{2}{*}{organizzazione dei paesi esportatori di petrolio arabo} \\
    \\
    \cmidrule(lr){2-3}
    & \multirow{3}{4em}{Context + \\ Question \\ (Source)} & Negli Stati Uniti, gli studiosi sostengono che esisteva già un accordo negoziato basato sull' uguaglianza tra le due parti prima del 1973. La possibilità che il Medio Oriente potesse diventare un altro confronto di superpotenza con l' URSS era più preoccupante per gli Stati Uniti che per il petrolio. Inoltre, i gruppi d' interesse e le agenzie governative più preoccupati per l' energia non erano all' altezza della dominanza di Kissinger. [...] \textit{Domanda: I gruppi d' interesse e le agenzie governative che si occupavano di energia non erano all' altezza di chi?} \\
    \cdashline{2-3}
    & \multirow{2}{4em}{Answer \\ (Generated)} & \multirow{2}{*}{kissinger} \\
    \\
    \midrule
    \multirow{3}{8em}{\bf{Question \\ Generation (SQuAD-IT QG)}} & ️\multirow{3}{4em}{Context + \\ Answer \\ (Source)} & In seguito all' evento di estinzione del Cretaceo-Paleogene, l' estinzione dei dinosauri e il clima umido possono aver permesso alla foresta pluviale tropicale di diffondersi in tutto il continente. Dal 66-34 Mya, la foresta pluviale si estendeva fino a sud fino a 45°. [...] Tuttavia, la foresta pluviale è riuscita ancora a prosperare durante questi periodi glaciali, consentendo la sopravvivenza e l' evoluzione di un' ampia varietà di specie. \textit{Risposta: l' estinzione dei dinosauri e il clima umido} \\
    \cdashline{2-3}
    & \multirow{2}{4em}{Question \\ (Generated)} & \multirow{2}{*}{quali fattori hanno permesso alla foresta pluviale tropicale di diffondersi in tutto il continente?} \\
    \\
    \cmidrule(lr){2-3}
    & \multirow{3}{4em}{Context + \\ Answer \\ (Source)} & Il primo europeo a percorrere il Rio delle Amazzoni fu Francisco de Orellana nel 1542. [...] Le Storie Unnatural Histories della BBC hanno presentato la prova che la foresta pluviale amazzonica, piuttosto che essere un deserto incontaminato, è stato modellato dall' uomo per almeno 11.000 anni attraverso pratiche come il giardinaggio forestale e la terra preta. \textit{Risposta: Francisco de Orellana} \\
    \cdashline{2-3}
    & \multirow{2}{4em}{Question \\ (Generated)} & \multirow{2}{*}{chi è stato il primo europeo a percorrere il rio delle amazzoni?} \\
    \\
    \bottomrule
    \end{tabular}
}
\vspace{-0.6mm}
\caption{Examples for the summarization and question answering/generation tasks from the respective test sets using the IT5 Base model.}
\label{tab:examples-1}
\end{table*}

\begin{table*}[!t]
\small 
\centering
\vspace{-2mm}
\scalebox{0.80}{
\def\arraystretch{1.5}
    \begin{tabular}{lp{6em}p{44em}}
    \toprule
    \bf{Task} & \bf{Field} & \bf{Examples} \\
    \midrule
    \multirow{3}{8em}{\bf{Headline Style \\ Transfer \\ (CHANGE-IT)}}  & ️\multirow{2}{6em}{La Repubblica \\ Article (Source)} &  ROMA. E adesso tocca all'arbitro. Da settimane ogni passo di Sergio Mattarella è orientato alla stabilità, ma lo schiaffo referendario sembra rendere impossibile un governo saldo con Matteo Renzi a Palazzo Chigi. Intendiamoci: per il Colle è sempre l'attuale capo dell'esecutivo a dover decidere se restare in sella. Non sarebbe certo il Capo dello Stato, insomma, a non favorire un "Renzi bis". Ma sono la politica, e lo stesso leader, ad essere già passati al piano B. [...] \\
    \cdashline{2-3}
    & \multirow{2}{6em}{Il Giornale \\ Headline (Gen.)} & \multirow{2}{*}{mattarella e il piano b: il premier al quirinale} \\
    \\
    \cmidrule(lr){2-3}
    & \multirow{3}{6em}{La Repubblica \\ Article (Source)} & CAMPOBASSO -  La prima regione "stellata" non c'è. Davide non ce l'ha fatta contro Golia, una sola lista non riesce a travolgerne 9 dello schieramento avversario. Così il M5s non replica fino in fondo l'onda del 4 marzo e il Molise va al centrodestra, e al suo candidato governatore, il sessantenne commercialista di Fi DonatoToma. [...] \\
    \cdashline{2-3}
    & \multirow{2}{6em}{Il Giornale \\ Headline (Gen.)} & \multirow{2}{*}{elezioni molise, il m5s non ce la fa} \\
    \\
    \midrule
    \multirow{3}{8em}{\bf{Headline \\ Generation \\ (HeadGen-IT)}} & ️\multirow{2}{6em}{News Article \\ (Source)} &  C'è chi perde la casa, tra una scossa e l'altra del terremoto, e vive un doppio dramma perché sfollato con un quattrozampe al seguito. Lo denuncia l'Enpa (Ente nazionale protezione animali) spiegando che, in alcune strutture di accoglienza destinate ai terremotati del centro Italia, gli animali domestici non sarebbero graditi. [...] Sono circa quindici i cani e i gatti tratti in salvo nelle ultime ore tra Norcia e le altre località del terremoto; la stima attuale è che 5mila tra cani e gatti abbiano seguito i loro padroni nelle tendopoli o negli alberghi messi a disposizione. [...]\\
    \cdashline{2-3}
    & \multirow{2}{6em}{Headline \\ (Generated)} & \multirow{2}{*}{animali al seguito rifiutati dagli albergatori appello dell'enpa: "anche loro sono sfollati"} \\
    \\
    \cmidrule(lr){2-3}
    & \multirow{3}{6em}{News Article \\ (Source)} & Parla della crisi del Venezuela, Papa Francesco: "Mi fa paura lo spargimento di sangue. E per questo chiedo di essere grandi a coloro che possono aiutare a risolvere il problema. Il problema della violenza mi atterrisce. E se hanno bisogno di aiuto che si mettano d\'accordo e lo chiedano". [...] Oltre ai pastori anche i cristiani, i cattolici ipocriti, che vanno tutte le domeniche a messa e poi non pagano la tredicesima, pagano in nero, sfruttano la gente, poi vanno ai Caraibi a fare le vacanze. `Ma io sono cattolico, vado tutte le domeniche a messa!'. Se tu fai questo dai una contro testimonianza. [...]\\
    \cdashline{2-3}
    & \multirow{2}{6em}{Headline \\ (Generated)} & \multirow{2}{*}{il papa: "evitare la violenza in venezuela". e sul lavoro: "ipocriti i cattolici che lo sfruttano"} \\
    \\
    \midrule
        \multirow{3}{8em}{\bf{Formal-to-Informal \\ Style Transfer \\ (XFORMAL F $\rightarrow$ I)}} & \multirow{2}{6em}{Formal \\ (Source)} & \multirow{2}{*}{evita di opprimerlo eccessivamente} \\
    \\
    \cdashline{2-3}
    & \multirow{2}{6em}{Informal \\ (Generated)} & \multirow{2}{*}{non opprimerlo troppo} \\
    \\
    \cmidrule(lr){2-3}
    & \multirow{2}{6em}{Formal \\ (Source)} & \multirow{2}{*}{esprimile ad alta voce l'amore che provi nei suoi confronti} \\
    \\
    \cdashline{2-3}
    & \multirow{2}{6em}{Informal \\ (Generated)} & \multirow{2}{*}{dille quanto ti piace} \\
    \\
    \cmidrule(lr){2-3}
    & \multirow{2}{6em}{Formal \\ (Source)} & sto facendo in modo di attivarmi per una richiesta ricevuta da lunaurora, dunque partecipo sicuramente alla chat. \\
    \cdashline{2-3}
    & \multirow{2}{6em}{Informal \\ (Generated)} & \multirow{2}{*}{io lo faccio per una richiesta lunaurora e partecipo sicuramente alla chat.} \\
    \\
    \midrule
    \multirow{3}{8em}{\bf{Informal-to-Formal \\ Style Transfer \\ (XFORMAL I $\rightarrow$ F)}} & \multirow{2}{6em}{Informal \\ (Source)} & \multirow{2}{*}{ciao lo so che nn centra nulla ma nn so propio come contattarti!!} \\
    \\
    \cdashline{2-3}
    & \multirow{2}{6em}{Formal \\ (Generated)} & \multirow{2}{*}{ciao, so che non c'entra nulla, ma non so come contattarti.} \\
    \\
    \cmidrule(lr){2-3}
    & \multirow{2}{6em}{Informal \\ (Source)} & concordo decisamente con verdina b xò nn t devi far vedere da lui!cm è 1 idea bellissima quella di verdina b! \\
    \cdashline{2-3}
    & \multirow{2}{6em}{Formal \\ (Generated)} & \multirow{2}{*}{concordo con verdina b, ma non devi farti vedere da lui.} \\
    \\
    \cmidrule(lr){2-3}
    & \multirow{2}{6em}{Informal \\ (Source)} & \multirow{2}{*}{meglio 1 pò di pancetta e tanta allegria ke 1 triste e insoddisfatta ragazza pelle e ossa!} \\
    \\
    \cdashline{2-3}
    & \multirow{2}{6em}{Formal \\ (Generated)} & \multirow{2}{*}{e' meglio avere un po' di pancetta e tanta allegria che una triste e insoddisfatta ragazza pelle e ossa.} \\
    \\
    \bottomrule
    \end{tabular}
}
\vspace{-0.6mm}
\caption{Examples of headline style transfer, headline generation and formality style transfer tasks from the respective test sets using the IT5 Base model.}
\label{tab:examples-2}
\end{table*}
\end{document}